\documentclass[submission,copyright,creativecommons]{eptcs}
\providecommand{\event}{ICLP 2023} 
\usepackage{iftex}

\ifpdf
  \usepackage{underscore}         
  \usepackage[T1]{fontenc}        
\else
  \usepackage{breakurl}           
\fi

\usepackage{amsfonts}
\usepackage{amssymb}
\usepackage{amsthm}
\usepackage{amsmath}
\usepackage{todonotes}
\usepackage{xcolor}
\usepackage[normalem]{ulem}

\newtheorem{example}{Example}

\def\Args{Args}
\def\Att{Att}

\newcommand{\vdashacc}{\ensuremath{\models}}

\newcommand{\abaf}{\ensuremath{\langle {\cal L}, \, {\cal R}, \, {\cal A},\, \overline{ \vrule height 5pt depth 3.5pt width 0pt \hskip0.5em\kern0.4em}\rangle}}
\def\contrary{\overline{ \vrule height 5pt depth 3.5pt width 0pt \hskip0.5em\kern0.4em}}
\newcommand{\A}{\mathcal{A}}
\newcommand{\LL}{\mathcal{L}}
\newcommand{\RR}{\mathcal{R}}
\newcommand{\RRl}{\RR_{learnt}}

\newcommand{\sABA}{\langle\RR,\A,\contrary\rangle}
\newcommand{\sABAp}{\langle \RR',\A',\contrary'\rangle}

\newcommand{\Ep}{\mathcal{E}^+}
\newcommand{\En}{\mathcal{E}^-}

\newcommand{\EE}{\langle\Ep,\En\rangle}

\newcommand{\argur}[3]{#1 \vdash_{#3} #2} 
\newcommand{\argu}[2]{#1 \vdash #2} 
\newcommand{\ruleset}{\ensuremath{R}}
\newcommand{\sent}{\ensuremath{s}}

\newcommand{\asm}{\ensuremath{a}}
\newcommand{\asmset}{\ensuremath{A}}

\title{ABA Learning via ASP}

\author{Emanuele De Angelis
\institute{IASI-CNR\\
Rome, Italy\email{emanuele.deangelis@iasi.cnr.it}\\
{0000-0002-7319-8439}}
\and
Maurizio Proietti
\institute{IASI-CNR\\ 
Rome, Italy
\email{maurizio.proietti@iasi.cnr.it}\\
{0000-0003-3835-4931}}
\and Francesca Toni 
\institute{Department of Computing\\ Imperial College London, UK
\email{ft@ic.ac.uk}\\
{0000-0001-8194-1459}}}

\def\titlerunning{ABA Learning via ASP}
\def\authorrunning{ De Angelis, Proietti, Toni}
\begin{document}
\maketitle

\begin{abstract}
Recently, ABA Learning has been proposed as a form of symbolic machine learning for drawing Assumption-Based Argumentation frameworks from background knowledge and positive and negative examples.  We propose a novel method for implementing ABA Learning using Answer Set Programming as a way to help guide Rote Learning and generalisation in ABA Learning.
\end{abstract}

\section{Introduction}
Recently, \emph{ABA Learning} has been proposed~\cite{MauFraILP-22-CoRR} as a methodology for learning 
Assumption-Based Argumentation (ABA) frameworks~\cite{ABA,ABAhandbook} from a background knowledge, in the form of an ABA framework, and positive and negative examples, in the form of sentences in the language of the background knowledge. The goal of ABA Learning is to build a larger ABA framework than the background knowledge from which  arguments for all positive examples can be ``accepted''  and no arguments for any of the negative examples can be ``accepted''. In this paper, for a specific form of ABA frameworks corresponding to logic programs~\cite{ABA}, we focus on a specific form of ``acceptance'', given by cautious (or sceptical) reasoning under the argumentation semantics of stable extensions~\cite{ABA,ABAhandbook}. We then leverage on the well known correspondence between stable extensions in the logic programming instance of ABA and answer set programs~\cite{GeL91} to outline a novel implementation strategy for the form of ABA Learning we consider, pointing out along the way restrictions on ABA Learning enabling the use of Answer Set Programming~(ASP).  

\paragraph{Related Work}

 Our strategy for ABA Learning differs from other works learning argumentation frameworks, e.g. \cite{DimopoulosK95,
 Tony-ArgML22}, 
 in that it learns a different type of argumentation frameworks 
and it uses ASP.
ABA can be seen as performing abductive reasoning (as assumptions are hypotheses open for debate).  
Other approaches combine  
abductive  and inductive learning~\cite{Ray09}, but they do not learn ABA frameworks.  
Some approaches learn abductive logic programs~\cite{InoueH00}, which rely upon assumptions, like ABA frameworks. 
A formal comparison with these methods is left for future work.
ABA captures several non-monotonic reasoning formalisms, thus ABA Learning is related to other methods learning non-monotonic formalisms. Some of these methods, e.g. \cite{InoueK97,Sakama00,ShakerinSG17}, do not make use of ASP. Some others, 
e.g. \cite{LawRB14,Sakama05,SakamaI09}, do. While our use of ASP to help  guide some aspects of ABA Learning (e.g. its Rote Learning transformation rule) is unique, a formal and empirical comparison with these methods  is left for future work.

\section{Background}
\label{sec:background}

\paragraph{ASP}
In this paper we use \emph{answer set programs} (ASPs)~\cite{GeL91
} consisting of rules of the form

\smallskip
$\mathtt{p ~\texttt{:-} 
~q_1,\dots, q_k, ~not~q_{k+1},\ldots, ~not~q_{n} 
}$
  \quad \quad or \quad \quad 
$\mathtt{  \texttt{:-} 
~q_1,\dots, q_k, ~not~q_{k+1},\ldots, ~not~q_{m} 
}$

\smallskip
\noindent
where ${\tt p }$, ${\tt q}_1$, $\dots$, ${\tt q_n}$,
${\tt q}_1$, $\dots$, ${\tt q_m}$ are atoms, 
$\mathtt{k}\geq 0$,
$\mathtt{n}\geq 0$,
$\mathtt{m}\geq 1$, 
and ${\tt not }$ denotes negation as failure.

Given any ASP program $P$, by $\textit{ans}(P)$,
called \textit{answer set} of $P$, we denote the set of ground atoms assigned to $P$ by the answer set semantics.
Let $\textit{ans}_1(P),\dots,\textit{ans}_l(P)$
be the answer sets of $P$, for $l \geq 1$ (if $l=0$, then $P$ is \emph{unsatisfiable}).
By 
${\mathcal C(P)}\!=\!\bigcap_i\textit{ans}_i(P)$, 
we denote the set of 
\emph{cautious} consequences of $P$
. 

\paragraph{ABA} 
An {\em ABA framework} (as originally proposed in \cite{ABA}, but presented here following  recent accounts in  \cite{ABAbook,ABAtutorial} and 
\cite{ABAhandbook}) is a tuple \abaf{}
such that

\vspace*{-0.2cm}
\begin{itemize}
\item 
$\langle \LL, \RR\rangle$ is a deductive system,
 where $\LL$ is a \emph{language} and $\RR$ is a set of
 \emph{(inference) rules} of the form $\sent_0 \leftarrow \sent_1,\ldots, \sent_m $ ($m \ge 0, \sent_i \in \LL$, for $1\leq i \leq m$); 

\item 
$\A$ $\subseteq$ $\LL$ is a (non-empty) 
set
of {\em assumptions};\footnote{The non-emptiness requirement
 can always be satisfied by including in $\A$ a \emph{bogus assumption}, with its own contrary, neither occurring elsewhere in the ABA framework. 
For 
conciseness, we will leave this assumption and its contrary implicit.} 

\item 
$\contrary$ is a total mapping from $\A$ into
 $\LL$, where $\overline{\asm}$ is the {\em contrary} of $\asm$, for $\asm
  \in \A$.
\end{itemize}
\vspace*{-0.2cm}
Given a rule $\sent_0 \gets \sent_1, \ldots,
\sent_m$, $\sent_0$ is 
the {\em head} 
and $\sent_1,\ldots, \sent_m$ 
is 
the {\em body}; 
if $m=0$ then the  body is said to be {\em empty} (represented as $\sent_0 \gets$ or  $\sent_0 \gets true$) and the rule is called a \emph{fact}.
If assumptions are not heads 
of rules then the ABA framework is called
{\em flat}. In this paper we focus on flat ABA frameworks.  
Elements of $\LL$ can be any sentences, but in this paper we  focus on (flat) ABA frameworks where $\LL$ is a set of ground atoms.
However, in the spirit of logic programming, we will use \emph{schemata} for rules, assumptions and contraries, using variables to represent compactly all instances over some underlying universe.


\begin{example}
    \label{ex:innocentABA}
The following is a flat ABA framework with $\LL$ a set of atoms.  
\vspace*{-0.2cm}
\begin{eqnarray*}
&\mathcal{R}=& \{ innocent(X) \leftarrow person(X), \, not\_guilty(X), \quad guilty(X) \leftarrow 
witness\_con(X,Y), \\
&& person(mary)\leftarrow ,  
\quad person(alex)\leftarrow, 
\quad witness\_con(mary,alex)\leftarrow\} \\
& \mathcal{L} = & \{innocent(X),  person(X), not\_guilty(X), guilty(X), witness\_con(X,Y) | X,Y \in \{mary, 
alex\} \}\\
& \mathcal{A} = & \{not\_guilty(mary)
\} \quad {\rm where } \quad \overline{not\_guilty(mary)}=guilty(mary)
.
\end{eqnarray*}
\end{example}
The semantics of flat ABA frameworks is given in terms of ``acceptable'' extensions, i.e.~sets of \emph{arguments} able to ``defend'' themselves against {\em attacks}, in some sense, as determined by the chosen semantics.
Intuitively, arguments are deductions of claims using rules and 
supported by assumptions, and  attacks are directed at the
assumptions in the support of arguments.  
For illustration, in the case of Example~\ref{ex:innocentABA},
there are, amongst others, arguments $\argur{\{not\_guilty(mary)\}}{innocent(mary)}{\{\rho_1,\phi_1\}}$ (with $\rho_1$ the first rule in $\RR$ and $\phi_1$ the  fact $person(mary)\leftarrow$ in $\RR$) and
$\argur{\emptyset}{guilty(mary)}{\{\rho_2,\phi_3\}}$ (with $\rho_2$ the second rule in $\RR$ and $\phi_3$ the  fact $witness\_con(mary,alex)\leftarrow$ in $\RR$), 
with the latter argument attacking the former. 

Given a flat ABA framework \abaf, 
let $\Args$ be the set of all arguments and $\Att=\{(\alpha,\beta) \in \Args \times \Args \mid \alpha$ attacks $\beta\}$
.
Then, the notion of ``acceptable'' extensions we  will focus on in this paper is as follows: $\Delta\subseteq \Args$ is a \emph{stable extension} iff (i) $\nexists \alpha,\beta \!\in \!\Delta$ such that $(\alpha,\beta) \!\in \!\Att$ (i.e. $\Delta$ is \emph{conflict-free}) and (ii) $\forall \beta \!\in \!\Args\setminus \Delta, \exists \alpha \!\in \!\Delta$ such that $(\alpha,\beta) \!\in \!\Att$ (i.e. $\Delta$ ``attacks'' all arguments it does not contain, thus pre-emptively ``defending'' itself against potential attacks).
%
We will consider 
the 
\emph{cautious} (a.k.a. \emph{sceptical}) consequences of (flat) ABA frameworks \abaf, i.e. the sets of sentences in $\LL$ that are claims of arguments in 
all
stable extensions for \abaf.
The 
cautious consequences of  the ABA framework in Example~\ref{ex:innocentABA} 
include $guilty(mary)$ and $innocent(alex)$.

Here we will work with ABA frameworks admitting at least one stable extension.
Also, without loss of generality, 
we will leave the language component of all ABA frameworks implicit,
and use, e.g., $\sABA $  to stand for $\abaf $ where $\LL$ is the set of all sentences in $\RR$, $\A$ and in the range of~$\contrary$.
We will also use $\sABA \vdashacc s$  to indicate 
that $s\in \LL$ is a cautious consequence  of $\sABA$
.

\section{Preliminaries: Cautious ABA Learning under 
Stable Extensions}
\label{sec:ABAlearning}

Here we recap the instance of the ABA Learning method proposed in \cite{MauFraILP-22-CoRR} that we focus on implementing using ASP in this paper, while stating restrictions on ABA Learning required by the implementation.

In \cite{MauFraILP-22-CoRR}, the \emph{background knowledge} is \emph{any} ABA framework $\sABA$.  Here, we assume that (i) it is restricted so that each assumptions occurs in the body of at most one rule schema in $\RR$, (ii)  for each non-ground  $\alpha(X)\in \A$ in the body of any 
$\rho\in \RR$, for $X$ a tuple of variables, for each variable $X'$ in $X$, there is at least one \emph{non-assumption} 
(in $\LL \setminus \A$) $p(Y)$ 
in the body of $\rho$ with $X'$ in $Y$, and (iii) each fact in $\RR$ is ground.
Restriction (i) is without loss of generality; the other two derive from  the use of schemata
.

In \cite{MauFraILP-22-CoRR}, \emph{positive/negative examples} are ground atoms of the form $p(c)$, for $p$ a predicate with arity $n \geq 0$ and $c$ a tuple of $n$ constants.
Here, we impose that examples are non-assumptions (in the background knowledge $\sABA$). 
So, for $\LL$ as in Example~\ref{ex:innocentABA},
 $not\_guilty$ cannot appear in 
 examples (but contraries
 , e.g. $guilty(mary)$, can be examples).
 The exclusion of assumptions from examples is derived from the flatness restriction. 
We also assume that for each example
$p(c)$ and $c'$ in $c$, 
$\exists\, q(d)\leftarrow \in \RR$ such that $c'$ is in $d$. We impose the same restriction on constants $c'$ anywhere 
in the background knowledge.

Given background knowledge $\sABA$, positive examples $\mathcal{E}^+$ and
negative examples $\mathcal{E}^-$
with $\mathcal{E}^+\cap \mathcal{E}^- =\emptyset$,
the \emph{goal of ABA Learning} is to 
construct (a flat ABA framework)
$\sABAp$ such that 
$\RR \subseteq \RR'$, 
$\A \subseteq \A'$,
and $\overline{\alpha}'=\overline{\alpha}$ for all $\alpha \in \A$, so that $\sABAp$ \textit{entails} 
$\EE$, that is
:

\noindent 
(\emph{Existence}) $\sABAp$ admits at least one stable extension,

\noindent
(\emph{Completeness}) for all $e \in \mathcal{E}^+$, $\sABAp \models e$, and

\noindent
(\emph{Consistency}) for all $e \in \mathcal{E}^-$, $\sABAp \not\models e$.

$\sABAp$ is called a \textit{solution} of the ABA Learning problem $(\sABA, \EE)$.
%
The second condition implies, when  $\mathcal{E}^+\neq \emptyset$,   that the set of cautious consequences  of a solution is non-empty.

In this paper we strive towards what we may call 
\emph{intensional solutions}, namely such  that $\RR' \setminus \RR$ comprises 
of   
\emph{intentional rules} (i.e. non-ground rule schemata), to avoid or limit ``lazy'' learning of  facts covering the positive examples alone and none of the negative examples, leading to poor generalisation
. 


\begin{example}
    \label{ex:innocentABAlearning}

Consider the background knowledge:
\begin{eqnarray*}
&\RR=& \{ innocent(X) \leftarrow \textit{defendant}(X), \, not\_guilty(X), \\
&& witness\_con(mary,alex) \leftarrow, \quad witness\_con(david,carol) \leftarrow, \quad witness\_con(john,carol) \leftarrow, \\
&& \textit{defendant}(mary)\leftarrow, \ \textit{defendant}(david)\leftarrow, \ \textit{defendant}(john)\leftarrow,  \ liar(alex) \leftarrow, \ away(bob)\leftarrow,\\
&& person(alex)\leftarrow, \quad  person(bob) \leftarrow, \quad  person(carol)\leftarrow, \\
&&\textit{person}(mary)\leftarrow, \quad \textit{person}(david)\leftarrow, \quad \textit{person}(john)\leftarrow\}\\
&\A=&\{not\_guilty(X) \mid X \in \{mary,david,john\}\} \quad {\rm where } \quad \overline{not\_guilty(X)}=guilty(X)
\end{eqnarray*}
and examples
$\mathcal E^+ =\{innocent(mary), \ innocent(bob)\}$,
$\mathcal E^-=\{innocent(david), \ innocent(john)\}$.
Then, solutions  of $(\sABA, \EE)$ include
ABA frameworks with $\RR_1'$ and $\RR_2'$
whereby

$\RR_1'\setminus \RR=\{
innocent(bob) \leftarrow, \quad guilty(david)\leftarrow, \quad guilty(john) \leftarrow \}$ 
and

$\RR_2'\setminus \RR=\{innocent(X) \leftarrow away(X), \quad  guilty(X) \leftarrow witness\_con(X,Y), \, person(Y), \, a(X,Y), $ 

\phantom{$\RR_2'\setminus \RR=\{$}
$c\_a(X,Y) \leftarrow \textit{defendant}(X), liar(Y) \}$. 
\\
The latter can be deemed to be intensional, whereas the former is not.
\end{example} 
In the remainder, as in \cite{MauFraILP-22-CoRR}, we will represent ground facts $p(t) \leftarrow $ as $p(X) \leftarrow  {X}={t}$. We will also use $vars(A)$ to denote the set of all variables occurring in assumption, rule or rule body  $A$.

\section{Learning ABA Frameworks via Transformation Rules and ASP Solving}
\label{subsec:transformation}

In order to learn ABA frameworks from  examples, we follow the approach based on \emph{transformation rules} presented in \cite{MauFraILP-22-CoRR}, but only consider a subset of those rules: \textit{Rote Learning}, \textit{Folding}, \textit{Assumption Introduction}, and (a special case of) \textit{Subsumption} 
(thus ignoring \textit{Equality Removal}).
 Some  rules (Folding and Subsumption) are borrowed from logic program transformation~\cite{PeP94
 }, while others (Rote Learning and Assumption Introduction) are specific for ABA.
Given an ABA framework $\sABA$, a transformation rule constructs a new ABA framework $\sABAp$ (in the remainder, we will mention explicitly only the modified components
).
The application of the transformation rules is guided by the \emph{ASP-ABALearn} strategy (see Figure~\ref{fig:strategy}),
a variant of the strategy in~\cite{MauFraILP-22-CoRR}
amenable to be implemented via an ASP solver, 
towards the goal of deriving an intensional solution of the given ABA Learning problem.  

\begin{figure}[h]
\hrule
\smallskip
\noindent
\textbf{Strategy} 
\textit{ASP-ABAlearn.} 
%
\textbf{Input:} An ABA Learning problem $(\sABA,\EE)$;

\quad $\textit{RoLe}(\sABA, \EE)$; \quad
$\textit{GEN}(\sABA, \EE)$.

\bigskip

\noindent
\textbf{Procedure} $\textit{RoLe}(\sABA, \EE)$.

\smallskip

$P := ASP^*(\sABA,\EE,\A)$;

\textbf{if}  $P$ is unsatisfiable 
\textbf{then} fail;

\textbf{for} all $\mathtt {c\_\alpha(t)} \in \mathcal C(P)\setminus \mathcal C(ASP(\sABA))$, where ${c\_\alpha(t)}= \overline {\alpha(t)}$ for some $\alpha(t) \in \A$ \textbf{do}

\quad apply \textit{Rote Learning} and get $\mathcal R := \mathcal R \cup \{ {c\_\alpha(X) \leftarrow X=t}\}$;

\textbf{for} all $p(u) \in \Ep$ such that $\mathtt {p(u)} \not\in\mathcal C(ASP(\sABA)) \ $ \textbf{do}

\quad apply  \textit{Rote Learning}  and get $\mathcal R := \mathcal R \cup \{ {p(X) \leftarrow X=u}\}$;

\textbf{if}  $\sABA$ does not entail $\EE$ 
\textbf{then} ~fail;

\bigskip

\noindent
\textbf{Procedure} $\textit{GEN}(\sABA, \EE)$.

\smallskip

\textbf{while} there exists a non-intensional rule $\rho_1\in \RRl$ \textbf{do} 

\smallskip

\quad // \textbf{Folding:}

\quad $\rho_2:=\textit{fold-all}(\rho_1)$;
\quad $\RR := (\RR\setminus \{\rho_1\}) \cup \{\rho_2\}$;

\smallskip

\quad // \textbf{Assumption Introduction:} 
%

\quad \textbf{if}  $\sABA$ does not entail $\EE$ \textbf{then}

\quad \quad let $\rho_2$ be of the form $H \leftarrow B$;
apply \textit{Assumption Introduction} and get 
$\rho_3$: $H \leftarrow B, \alpha(X)$

\quad \quad where $\alpha$ is a new predicate symbol and $X=vars(H \leftarrow B)$;

\quad \quad $\RR := (\RR\setminus\{\rho_2\}) \cup \{\rho_3\}$;
\quad $\A := \A\cup \{\alpha({X})\}$, with:
$\overline{\alpha({X})}=c\_\alpha({X})$;

\smallskip

\quad \quad // \textbf{Rote Learning:}



\quad \quad \textbf{for} all $\mathtt {c\_\alpha(t)} \in \mathcal C(ASP^+(\sABA,\EE,\{\alpha({X})\}))$ \textbf{do}

\quad \quad \quad apply \textit{Rote Learning} and get $\mathcal R := \mathcal R \cup \{ {c\_\alpha(X) \leftarrow X=t}\}$;

\quad \quad \textbf{if}  $\sABA$ does not entail $\EE$ \textbf{then} ~fail;
\smallskip

\quad \quad // \textbf{Subsumption:}

\quad \quad \textbf{for} all $\rho$: $p(X) \leftarrow X=t \in \RRl$ \textbf{do}

\quad \quad \quad \textbf{if} $\mathtt{p(t)} \in \mathcal C(ASP(\langle\RR\setminus \{\rho\},\A,\contrary\rangle)$ \textbf{then} ~apply  \textit{Subsumption} and delete $\rho$: $\RR:=\RR\setminus \{\rho\}$.

    \caption{\textit{ASP-ABAlearn} strategy. By $\RRl$ we denote the subset of the rules in 
    $\RR$ that do not belong to the original background knowledge.
    The function $\textit{fold-all}$, given a non-intensional rule $\rho_1$, returns an intensional rule $\rho_2$ obtained by applying once or more times the Folding rule to $\rho_1$ using rules in $\mathcal R$.  }
    \label{fig:strategy}
\vspace{2pt}
\hrule
\end{figure}

The \emph{ASP-ABAlearn} strategy is the composition of two procedures: (1) $RoLe$, which has the goal of adding suitable facts to the initial background knowledge $\sABA$ so that the new ABA framework $\sABAp$ is a (non-intensional) solution of the learning problem $(\sABA, \EE)$ given in input,  and (2) $GEN$, which has the objective of transforming $\sABAp$ into an intensional solution.
In general, it is not obvious which fact should be added to $\RR$ in $RoLe$ and how to generalise a non-intensional solution to obtain an intentional one in $GEN$
. For these purposes, we use various encodings into ASP defined in Fig.\ref{fig:rewrite} 
to obtain the following sets of ASP rules:
\begin{itemize}
    \item The set of ASP rules at points (a) and (b.1) of Fig.~\ref{fig:rewrite} is denoted by $ASP(\sABA)$. For a claim $s\in\LL$, we can check that $\sABA \vdashacc s$  (i.e., $s$ is a cautious consequence  of $\sABA$ under stable extensions) by checking that $\texttt{s} \in \mathcal C(ASP(\sABA))$.\vspace{-3pt}
    
    \item The set of ASP rules at points (a)--(d) is denoted $ASP^+(\sABA, \EE, \mathcal K)$. By computing $\mathcal C(ASP^+(\sABA, \EE, \mathcal K))$ we generate facts, if at all possible, for contraries of assumptions in $\mathcal K$ that, when added to $\RR$, enable the entailment of the examples in $\EE$.\vspace{-3pt}
    
    \item The set of ASP rules at points (a)--(e) is denoted $ASP^*(\sABA, \EE, \mathcal K)$.  It can be used similarly to $ASP^+(\sABA, \EE, \mathcal K)$, but also generates facts for positive examples that cannot be obtained by $ASP^+(\sABA, \EE, \mathcal K)$.
\end{itemize}

\begin{figure}[t]
\hrule
\smallskip
\begin{itemize}
    \item[(a)] Each rule in $\mathcal R$ is rewritten in the ASP syntax 
    (e.g. $innocent(X) \!\leftarrow  \!\textit{defendant}(X), not\_guilty(X)$ in Ex.~\ref{ex:innocentABAlearning} becomes ${\tt innocent(X)~\texttt{:-}~defendant(X),~not\_guilty(X).}$). In the following, we use the teletype font for the ASP translation of the rules.

\item[(b.1)] Each $\alpha_i \in \A$ occurring in the body of $H \leftarrow B, \alpha_1, \ldots, \alpha_n \in \RR$ is encoded as the following ASP rule,
where
$\mathtt c\_\alpha_i$ is an ASP atom encoding $\overline{\alpha_i}$, 
and $vars(\mathtt{\alpha_i}) \subseteq \mathtt{X}$:

\quad $\mathtt {\alpha_i~\texttt{:-}~dom(X),~not~c\_\alpha_i.}$

%

\item[(b.2)] Each $\alpha_i\in\mathcal K$ occurring in the body of $H \leftarrow B, \alpha_1, \ldots, \alpha_n \in \RR$ is encoded as the following pair of ASP rules: 

\quad $\mathtt {c\_\alpha_i~\texttt{:-}~dom(X),~not~\alpha_i.}$
\quad \quad \quad $\mathtt {\texttt{:-}~\alpha_i, ~c\_\alpha_i.}$


\item[(c)] Each $e \in \Ep$ is encoded as the ASP rule 
\texttt{:- not e.}

\item[(d)] Each $e\in \En$ is encoded as the ASP rule 
\texttt{:- e.}
    
\item[(e)] Each atom $p(X)\in \Ep\cup\En$ 
such that $p(X)$ is not the contrary of any assumption in~$\A$ is additionally encoded as the following triple of ASP rules, where $\mathtt {neg\_p}$ is a new predicate name
:

\quad $\mathtt {p(X)~\texttt{:-}~dom(X),~not~neg\_p(X).}$
\quad\quad$\mathtt {neg\_p(X)~\texttt{:-}~dom(X),~not~p(X).}$
\quad\quad$\mathtt {\texttt{:-}~p(X), ~neg\_p(X).}$

    
\end{itemize}

\vspace{-3mm}
\caption{\label{fig:rewrite}
ASP-encodings for a given  ABA Learning problem $(\sABA,\EE)$ and  a set $\mathcal K\subseteq \A$ of assumptions. 
Here, $\mathtt {dom}$  is chosen so that $\mathtt {dom(X)}$ holds for all $ \mathtt X$  encoding tuples of constants of $\LL$.
Note that, 
without loss of generality,
 in (b.1)-(b.2) we can assume $vars(\alpha_1, \!\ldots, \!\alpha_n) \!\subseteq \! vars(B)$. So, we could replace $\mathtt {dom(X)}$ by any subset of $B$ that contains all variables of $\mathtt{\alpha_i}$ and {\tt dom} may already occur in $\sABA$ 
    (this is an optimisation, as fewer ground instances of the rule may be given by the ASP solver). 
Also, in (e) {\tt dom} may already occur in $\sABA$. 
}
\vspace{2pt}
\hrule
\end{figure}

The $RoLe$ procedure repeatedly applies 
Rote Learning, which adds a 
fact $\rho :  p({X}) \leftarrow  {X}={t}$, where $t$ is a tuple of constants, to $\RR$ (thus, $\RR'\!=\!\RR\cup \{\rho\}$).
We illustrate with the \textit{innocent} running example.

\clearpage
\begin{example}
\label{ex:R1}
Let us consider the learning problem $(\sABA,\EE)$ of Example~\ref{ex:innocentABAlearning}.
In this case, $ASP^*(\sABA,\EE,\A)$ consists of $\RR$  rewritten in ASP syntax
and of the following ASP rules
:

\begin{small} 
\begin{verbatim}
not_guilty(X) :- person(X), not guilty(X).            :- not innocent(mary).
guilty(X) :- person(X), not not_guilty(X).            :- not innocent(bob).
:- not_guilty(X), guilty(X).                          :- innocent(john).
innocent(X) :- person(X), not neg_innocent(X).        :- innocent(david). 
neg_innocent(X) :- person(X), not innocent(X).        
:- innocent(X), neg_innocent(X). 
\end{verbatim}
\end{small}

\noindent
In this example, 
$\mathcal{C}(ASP^*(\sABA,\EE,\A))\setminus \mathcal{C}(ASP(\sABA))$ includes \begin{small}{\tt innocent(bob)}, {\tt guilty(david)}, {\tt guilty(john)} \end{small}, and 
\noindent
by Rote Learning, we obtain:

\smallskip

$\RR_1 = \RR \cup\{innocent(X) \leftarrow X=bob,\
guilty(X) \leftarrow X=david,\
guilty(X) \leftarrow X=john\}$

\smallskip

\noindent
$\langle \RR_1, \A, \contrary\rangle$ is a (non-intensional, due to the added ground facts) solution of our ABA Learning problem.
\end{example}

\smallskip
Now, the  {\em ASP-ABAlearn} learning strategy proceeds by applying the $GEN$ procedure, for transforming a non-intensional rule into an intensional one.
First,  $GEN$ applies (once or more times) Folding, which, given rules
$\rho_1$: $H \leftarrow Eqs_1, B_1, B_2$ and
$\rho_2$: $K \leftarrow Eqs_1, Eqs_2, B_1$
in $\RR$, replaces $\rho_1$ by 
$\rho_3$: $H \leftarrow Eqs_2, K, B_2$ 
%
%
(hence, $\RR'=(\RR\setminus \{\rho_1\})\cup\{\rho_3\}$).
%
Folding is a form of {\em inverse resolution}~\cite{Muggleton1995}, which generalises a rule by replacing some atoms in its body with their `consequence' using a rule in~$\RR$.

\smallskip

\begin{example}
\label{ex:R2}
%
%

By applying Folding to 
$innocent(X) \leftarrow X\!=\!bob$ and $guilty(X) \leftarrow X=david$, using \linebreak 
$away(X) \leftarrow X\!=\!bob$, $witness\_con(X,Y) \leftarrow X\!=\!david,\ Y\!=\!carol$, and $person(Y) \leftarrow Y\!=\!carol$ in $\RR_1$, we get: 

\smallskip

$\RR_2 \!=\!   \RR \cup \{\,innocent(X) \leftarrow away(X), \ guilty(X) \leftarrow witness\_con(X,Y), \, person(Y),$

~~~~~~~~~~~~~~~~~~$ \ guilty(X) \leftarrow X\!=\! john\}$.

\smallskip

\noindent
The ABA framework $\langle \RR_2, \A, \contrary\rangle$, though, is not a solution, as it does no longer entail $innocent(mary)$. 
\end{example}

If the effect of Folding a rule is that the resulting ABA framework is no longer a solution of the given learning problem, $GEN$ applies Assumption Introduction and Rote Learning with the goal of deriving a new ABA framework which is again a (non-intensional) solution. Assumption Introduction replaces a rule
$\rho_1: H \leftarrow B$ in $\RR$ by
$\rho_2: H \leftarrow B, \alpha({X})$,
where $X=\textit{vars}(H) \cup \textit{vars}(B)$ and $\alpha({X})$ is a new assumption with contrary $c\_\alpha({X})$ 
(thus, $\RR'=(\RR\setminus \{\rho_1\})\cup\{\rho_2\}$, $\mathcal{A}'= \mathcal{A} \cup \{\alpha({X})\}$, $\overline{\alpha({X})}'=c\_\alpha({X})$,
and $\overline{\beta}'=\overline{\beta}$ for all $\beta \in \A$). New facts for $c\_\alpha({X})$ are learnt by Rote Learning by using $ASP^+(\sABAp,\EE,\{\alpha(X)\}$. The facts for $c\_\alpha({X})$ can be seen as the \textit{exceptions} to the  \textit{defeasible} rule $\rho_2$.

\smallskip

\begin{example}
\label{ex:R3}
By Assumption Introduction, we get: 

\smallskip

$\RR_3 =   \RR \cup \{innocent(X) \leftarrow away(X), \quad  guilty(X) \leftarrow witness\_con(X,Y), \, person(Y), \, a(X,Y), $

~~~~~~~~~~~~~~~~~~~$guilty(X) \leftarrow X= john\}$ 

\smallskip
\noindent
with $\A'=\A\cup\{a(X,Y) \mid X,Y \in \{alex,$ $bob,carol,david,john,mary\}\}$ and $\overline{a(X,Y)}'\!\!=\!c\_a(X,Y)$. To determine the facts for $c\_a(X,Y)$, we use $ASP^+(\langle \RR_3,\A,\contrary\rangle,\EE,\{a(X,Y)\})$  (we omit the encoding of the background  knowledge~$\mathcal R$):


\begin{small}
\begin{verbatim}
innocent(X) :- away(X).                                    :- not innocent(mary).
guilty(X) :- X=john.                                       :- not innocent(bob).               
guilty(X) :- witness_con(X,Y), person(Y), alpha1(X,Y).     :- innocent(david).
alpha1(X,Y) :- witness_con(X,Y), not c_alpha1(X,Y).        :- innocent(john).
c_alpha1(X,Y) :- witness_con(X,Y), not alpha1(X,Y).     
:- alpha1(X,Y), c_alpha1(X,Y). 
\end{verbatim}
\end{small}

\smallskip

$\mathcal C(ASP^+(\langle \RR_3,\A,\contrary\rangle,\EE,\{a(X,Y)\}))$ contains the atom $\mathtt{c\_alpha1(mary,alex)}$ and thus, by Rote Learning, we obtain again a (non-intensional) solution, by adding a fact for predicate $c\_a(X,Y)$:

\smallskip

$\RR_4=\RR_3 \cup \{c\_a(X,Y) \leftarrow X=mary, \, Y=alex\}$

\end{example}

\smallskip

\noindent
$GEN$ proceeds by applying the Subsumption rule, which gets rid of redundant facts. Indeed, suppose that $\RR$ contains 
$\rho: p(X) \leftarrow X=t$
and let $\RR'=\RR\setminus \{\rho\}$. If $\sABAp \models p(t)$, then, by Subsumption,  $\rho$ can be deleted from $\RR$. Subsumption is applicable if $\mathtt{p(t)} \in \mathcal C(ASP(\sABAp))$.

\smallskip

\begin{example}
\label{ex:R4}
The rule 
$\rho=guilty(X) \!\!\leftarrow \!\!X\!\!=\!\! john$ can be deleted
, as $\mathtt{guilty(john)} \!\in\! \mathcal C(ASP(\langle \RR_4\!\setminus\!\{
\rho\}))$.
\end{example}

\noindent
\textit{ASP-ABAlearn} halts when $GEN$ generates no new contrary, as Folding yields an intensional solution.

\begin{example}
\label{ex:final}
By two final 
applications of the 
Folding rule, $GEN$ gets:

\smallskip

$\RR_5=\RR \cup \{\,innocent(X) \leftarrow away(X), \quad guilty(X) \leftarrow witness\_con(X,Y), \, person(Y), \, a(X,Y), \quad $

~~~~~~~~~~~~~~~~~~~~$c\_a(X,Y) \leftarrow \textit{defendant}(X), liar(Y)\}$

\smallskip

\noindent
Now, $\langle \RR_5, \A', \contrary'\rangle$ is an intensional solution of the given learning problem.
\end{example}

\section{Discussion and Conclusion}
\label{sec:concl}

We have revisited a strategy recently proposed for learning ABA frameworks based on transformation rules~\cite{MauFraILP-22-CoRR}, and we have shown that, in the case of the stable extension semantics, many of the reasoning tasks used by that strategy can be implemented through an ASP solver. 

A proof-of-concept implementation of our \textit{ASP-ABAlearn} strategy 
is ongoing
using  
SWI-Prolog 
(v. 9.0.4)
and  the Clingo ASP solver
(v. 5.6.2). 
It consists of two Prolog modules implementing 
\textit{RoLe} 
and \textit{GEN} 
and two further modules implementing 
(i) the 
$ASP$, $ASP^+$,
and $ASP^*$  encodings, 
and 
(ii) the 
API to invoke Clingo from SWI-Prolog 
and collect the cautious consequences to be used by \textit{RoLe} and \textit{GEN}. 
%
The most critical issue 
for implementing $GEN$ is that the application of Folding is non-deterministic, as there may be different choices for the rules to be used for applying that transformation. 
Currently, we simply make use of a bound to limit the number of alternatives. 
The design of more sophisticated mechanisms to control Folding, e.g., based on the notion of \textit{information gain}\cite{ShakerinSG17}, is left as future work.

In addition to refining the implementation
, we also plan to perform an experimental comparison to non-monotonic ILP systems (such as Fold~\cite{ShakerinSG17} and ILASP~\cite{LawRB14}).
On the theoretical side, further work is needed to investigate conditions under which  \textit{ASP-ABAlearn}  is complete, in the sense that it terminates and finds a solution 
if it exists. 
A simple way of guaranteeing termination is based on a mechanism for avoiding the generation of contraries that are ``equivalent'' to previously generated ones. 
However, the solution obtained in this way is not guaranteed to be intensional.


\section*{Acknowledgments}

We thank support from the 
Royal Society, UK (IEC\textbackslash R2\textbackslash 222045 - International Exchanges 2022
)
.
De Angelis and Proietti are members of the INDAM-GNCS research group; they were partially supported by the PNRR MUR project PE0000013-FAIR, Italy.
Toni was partially funded by the 
ERC under
the EU’s Horizon 2020 research and innovation programme (grant agreement 
No. 101020934) and 
by 
J.P. Morgan and  the
Royal Academy of Engineering, UK, under the Research Chairs and Senior Research Fellowships scheme.

\end{document}